\documentclass[conference]{IEEEtran}

\IEEEoverridecommandlockouts
\usepackage{stfloats}
\usepackage{graphicx}
\usepackage{amsmath}
\usepackage[justification=centering]{caption}
\usepackage{url}
\usepackage{color}
\def\BibTeX{{\rm B\kern-.05em{\sc i\kern-.025em b}\kern-.08em
    T\kern-.1667em\lower.7ex\hbox{E}\kern-.125emX}}

\usepackage{bbding}
\usepackage{amsthm}

\usepackage{cite}

\usepackage{diagbox}
\usepackage{multirow}
\usepackage{subfigure}

\usepackage[misc]{ifsym}

\makeatletter
\def\ps@IEEEtitlepagestyle{%
    \def\@oddfoot{\mycopyrightnotice}%
    \def\@evenfoot{}%
}

\def\mycopyrightnotice{%
    {\footnotesize DOI reference number: 10.18293/SEKE2021-085 \hfill}
    \gdef\mycopyrightnotice{}%
}

\begin{document}

\title{ Conv-Reluplex : A Verification Framework For Convolution Neural Networks  }

% \author{\IEEEauthorblockN{1\textsuperscript{st} Zishan Li }
% \IEEEauthorblockA{\textit{School of Software Engineering,} \\
% \textit{Tongji University,}\\
% Shanghai, China \\
% lizishan@tongji.edu.cn}
% \and
% \IEEEauthorblockN{2\textsuperscript{nd} Jin Xu}
% \IEEEauthorblockA{\textit{School of Software Engineering,} \\
% \textit{Tongji University,}\\
% Shanghai, China \\
% slhjin@tongji.edu.cn}
% \and
% \IEEEauthorblockN{3\textsuperscript{rd} Bowen Du}
% \IEEEauthorblockA{\textit{Department of Computer Science} \\
% \textit{University of Warwick}\\
% Coventry, United Kingdom \\
% B.Du@warwick.ac.uk}
% \and
% \IEEEauthorblockN{4\textsuperscript{th} Miaomiao Zhang}
% \IEEEauthorblockA{\textit{School of Software Engineering,} \\
% \textit{Tongji University,}\\
% Shanghai, China \\
% miaomiao@tongji.edu.cn}
% }

\author{
\IEEEauthorblockN{
Jin Xu\IEEEauthorrefmark{1},
Zishan Li\IEEEauthorrefmark{1},
Miaomiao Zhang\IEEEauthorrefmark{1}$^{(\textrm{\Letter})}$,
Bowen Du\IEEEauthorrefmark{2}$^{(\textrm{\Letter})}$ 
}

\IEEEauthorblockA{\IEEEauthorrefmark{1}School of Software Engineering, Tongji University, Shanghai, China}
\IEEEauthorblockA{\IEEEauthorrefmark{2}Department of Computer Science, University of Warwick, Coventry, United Kingdom}
\IEEEauthorblockA{Email:  miaomiao@tongji.edu.cn; B.Du@warwick.ac.uk}

\thanks{
 $^{\textrm{\Letter}}$ Corresponding author}
 }

\maketitle

\begin{abstract}
In recent years, machine learning  has demonstrated impressive performance in many real-world tasks, especially in computer vision and natural language processing. However, to apply them in safety-critical systems one needs formal guarantees on the neural network outputs. The Reluplex tool is proposed to verify the safety of deep neural networks (DNNs), and in case the DNN fails to give a correct output, can generate adversarial examples.
%There are a lot of neural networking employing Leaky ReLU as their non-linear activation function. Therefore,
Since the tool can only handle DNNs, it is necessary to extend the tool to process image data. Therefore, in this paper,  we propose the Conv-Reluplex framework, which is designed to verify the convolutional layer and pooling layer in convolutional neural networks(CNNs), and  generate adversarial examples when classification is misguided. We conduct several experiments  on MNIST to evaluate our approaches. The results show that the original CNN is improved using the adversarial examples generated by our tool, and the precision of classification can be increased significantly.

\end{abstract}


\section{Introduction}
In recent years, machine learning \cite{goodfellow2016deep}\cite{riesenhuber1999hierarchical} has been widely used in various fields, such as image recognition \cite{krizhevsky2012imagenet}, speech recognition \cite{graves2013speech} and autonomous vehicles \cite{zhao2017deep}. Neural networks are trained over a finite training set and are expected to generalize, i.e., to behave correctly for previously-unseen inputs. However,  Szegedy et al. discovered that the input-output mapping learned by neural networks is discontinuous to a large extent \cite{szegedy2013intriguing}. It turns out that perturbed inputs similar to a correctly classified input could  be misclassified by deep learning models with high confidence,  which are generally called adversarial examples \cite{szegedy2013intriguing}. %capable of attacking a safety-critical system.
 %As  DNNs can react in unexpected and incorrect ways to even slight perturbations of their inputs,
%This unexpected behavior of DNNs is likely to result in unsafe systems, or re￾strict the usage of DNNs in safety-critical applications.
There is an urgent need for methods that can provide formal guarantees about neural networks behavior. Unfortunately, manual reasoning about large neural networks is impossible, as their structure renders them incomprehensible to humans. Automatic verification techniques are thus  needed.

Verification of neural networks is difficult as it is experimentally beyond the reach of general-purpose tools such as linear program (LP)
solvers or existing satisfiability modulo theories (SMT) solvers \cite{pulina2012challenging, bastani2016measuring, huang2017safety}.
In \cite{pulina2010abstraction}, the authors propose a method to verify  Multi-layer Perceptron (MLP) \cite{rumelhart1985learning} with sigmoid activation function. They point out the difficulty in scaling-up this technique, i.e., only able to tackle small networks with at most 20 hidden nodes \cite{pulina2012challenging}. In \cite{huang2017safety}, the authors propose an approach for verifying the local adversarial robustness of  DNNs based on a systematic exploration of a region. The verification process is still exponential in the number of features.  Katz et al propose an algorithm called Reluplex, which is efficient to verify DNNs with ReLU activation functions \cite{katz2017reluplex}. This is achi eved by leveraging the piecewise linear nature of ReLUs and attempting to gradually satisfy the constraints that they impose as the algorithm searches for a feasible solution. We call the algorithm Reluplex, for “ReLU with Simplex”.  Compared with \cite{huang2017safety} and \cite{pulina2010abstraction}, Reluplex can handle larger deep neural networks and guarantee that there are no irregularities hiding between the discrete points.

Reluplex mainly focuses on DNNs. To support image processing, it is  necessary to extend this tool with CNNs to make it more practical. To this end, we present a Conv-Reluplex framework on the basis of Reluplex.  The extension is able to verify  network robustness, that is, if Conv-Reluplex  finds that the adversarial robustness is not satisfied, a corresponding counter-example (adversarial example) will be generated in the form of an image type which shows an abnormal classification with respect to the network. So, in the Conv-Reluplex framework, the adversarial examples generation is conducted  during verification process, not alike those in the existing methods \cite{goodfellow2014explaining, moosavi2016deepfool, brendel2017decision}. As 
have been shown in \cite{pulina2010abstraction}\cite{moosavi2016deepfool}\cite{katz2017towards}\cite{Vanderbei1998Linear}, adversarial training can
improve the robustness of models. We thus make some tricks
in the tool to produce a large number of adversarial examples,
not just one, for our later training of CNNs. The experimental
results show that appropriate adversarial training is helpful to
enhance the adversarial robustness of the CNNs.

So, based on the current tool---Reluplex, we give the Conv-Reluplex framework to verify CNNs, meanwhile generate adversarial examples. We also have implemented the framework 
 \cite{dantzig2016linear} and conducted some experiments. In the following, we begin with some background on CNNs, and Reluplex in Section 2. In Section 3, we present Conv-Reluplex verification framework, also with an emphasis on generation of adversarial examples, followed by exprerimental results and analysis in Section 4.
% Related work is discussed in Section 6.
We conclude the paper in the last section.

%=================================================================================

\section{Preliminaries}
We first recall some definitions of CNNs, Adversarial robustness and Reluplex algorithm.

\subsection{Convolutional Neural Networks}

A convolutional neural network\cite{cnndefinition} is comprised of one or more convolutional layers (often with a subsampling step) and then followed by one or more fully connected layers as in a standard multilayer neural network. 
%An example of CNNs is shown in Fig. \ref{cnn_arch}. 
The architecture of CNNs is designed to take advantage of the 2D structure of an input image (or other 2D input such as a speech signal). This is achieved with local connections and tied weights followed by some form of pooling which results in translation invariant features. Another benefit of CNNs is that they are easier to train and have many fewer parameters than fully connected networks with the same number of hidden units.

%\begin{figure}[htpb]
%  \centering
%  \includegraphics[width=9cm]{figure/cnn_arch}
%  \caption{The architecture of CNN}
%  \label{cnn_arch}
%\end{figure}

\subsection{Adversarial Robustness}
Adversarial robustness \cite{katz2017reluplex} is a safety property, which measures the resilience of a neural network against the inputs with perturbation. When the input is \(x_{0}\), the output of network can be denoted as $f({x_0})$. A neural network is \(\delta\)-locally-robust at point \(x_{0}\) iff
\begin{equation*}
\forall x, \quad\left\|x-x_{0}\right\| \leq \delta \Rightarrow f(x)=f\left(x_{0}\right)
\end{equation*}
Intuitively, the above formula states that for input  \(x\) that is very close to  \(x_{0}\), the network assigns to  \(x\) the same label that it assigns to  \(x_{0}\); “local” thus refers to a local neighborhood around  \(x_{0}\). Larger values of \(\delta\) imply larger neighborhoods, and hence better robustness\cite{katz2017towards}.

\subsection{Reluplex Algorithm}
Reluplex is an SMT solver for a theory of linear real arithmetic with ReLU constraints. %; a novel decision algorithm for solving queries on deep neural networks with ReLU activation function.
The technique is based on extending the Simplex algorithm \cite{Vanderbei1998Linear}\cite{dantzig2016linear} to support the non-convex ReLUs in a way that allows their inputs and outputs to be temporarily inconsistent which will then be fixed as the algorithm progresses. In other words, DNNs and their properties can be directly encoded as conjunctions of linear formulas and ReLU constraints. Here,  a ReLU constraint satisfies  $x^f=\max(0, x^b)$, where $x^f$ and $x^b$ stand for the connection information of the nodes. To guarantee termination, some ReLU connections may need to be split upon. However, in many cases this is not required, resulting in a practically efficient
solution. The details  that are crucial to performance and
scalability, such as the use of floating-point arithmetic, bound derivation for
ReLU variables, and conflict analysis, are discussed in \cite{katz2017reluplex}. The  success in verifying properties of the ACAS Xu networks \cite{julian2018deep} indicates that the technique holds  potential for verifying real-world DNNs.

\section{a verification framework for CNNs by Conv-Reluplex}
%The Reluplex is able to verify and improve the adversarial robustness of networks. If the adversarial robustness is not satisfied for a classification task, there must exist an $x'$ ,  $x'\le \x\pm \delta$, with $x'$ belonging to a label different with that of $x$. We call $x'$ a counter-example, which shows the violation of the adversarial robustness property. As the tool is not able to directly verify the CNNs and generate their corresponding counter-examples,  we hereby propose the Conv-Reluplex framework to boost its ability on CNNs.

The Reluplex is able to verify and improve the adversarial robustness of networks. If the adversarial robustness is not satisfied for a classification task, there must exist an \(x^{\prime}\), \(\left\|x^{\prime}-x\right\| \leq \delta\),  with \(x^{\prime}\) belonging to a label different from that of \(x\). We call \(x^{\prime}\) a counter-example, which shows the violation of the adversarial robustness property. As the tool is not able to directly verify the CNNs and generate their corresponding counter-examples,  we hereby propose the Conv-Reluplex framework to boost its ability on CNNs.

 \begin{figure*}[htbp]
  \centering
  \includegraphics[scale=0.6]{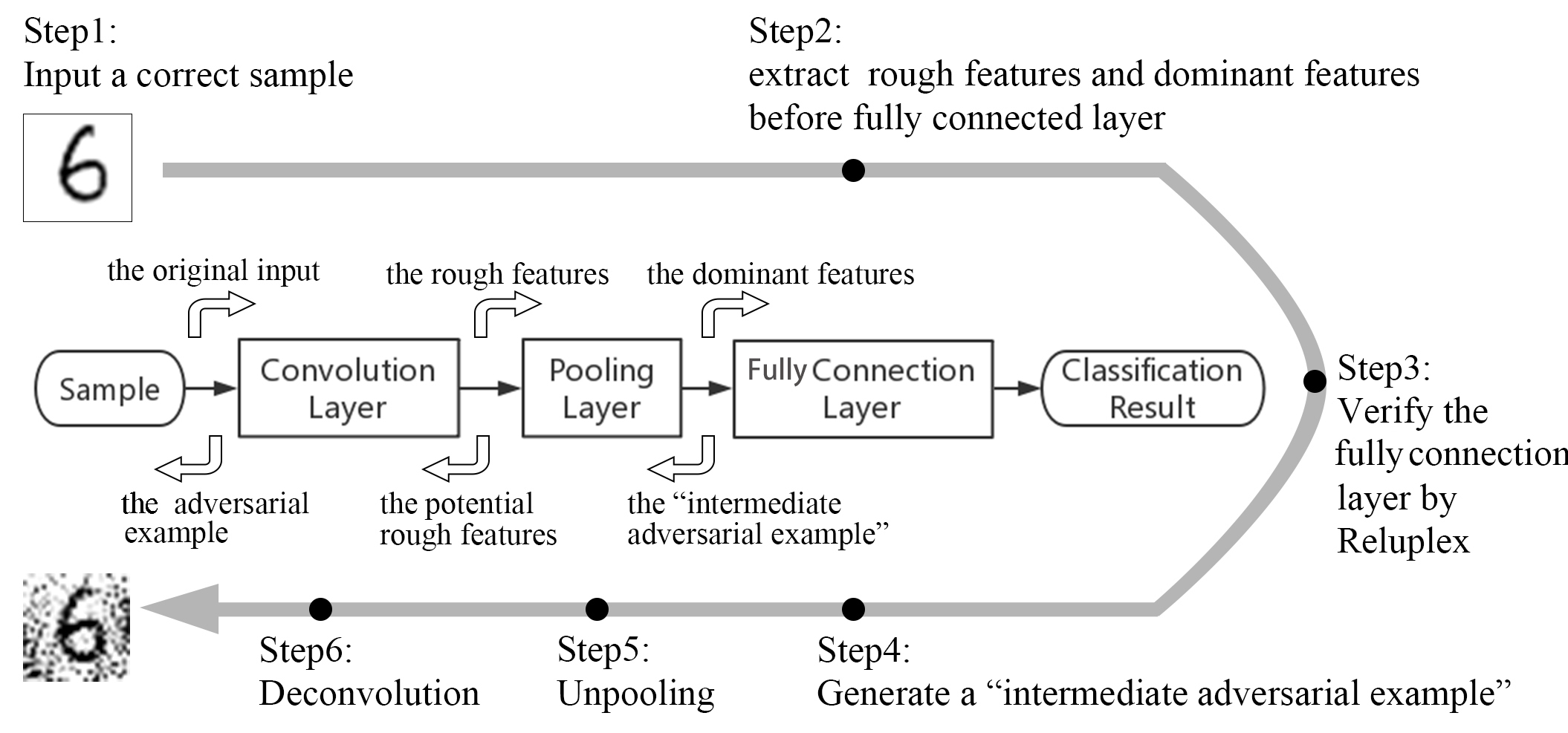}
  \caption{Main steps of the Conv-Reluplex framework}
  \label{Figure 4.1}
\end{figure*}

%\subsection{The main idea of Conv-Reluplex framework}
CNNs generally consist of three kinds of layers,  convolutional layers,  pooling layers and fully connected layers. The convolutional layer extracts rough features from the original image, the pooling layer is a form of non-linear down-sampling to generate dominant features, and the fully connected layer employs these dominant features for classification. The fully connected layer closely resembles the basic component of DNNs, so the Reluplex can be directly applied to process this layer. However, to tackle convolutional and pooling layers in the CNNs, in particular for the purpose of adversarial example generation, we propose a reverse calculation algorithm. % is proposed to pass the counter-examples, which generated by Reluplex, back through the pooling layer and the convolutional layer, and the counter-examples of CNNs are found.

As illustrated in Fig. \ref{Figure 4.1}, the designed Conv-Reluplex framework consists of Reluplex and a reverse calculation algorithm. %After the CNN is trained %(it performs well based on the training datasets).
Given a trained CNN classifier, there are six steps included in the framework.

% \begin{figure}[hpb]
%   \centering
%   \includegraphics[width=9cm]{figure/Figure_4_1_Main_steps}
%   \caption{Main steps of the Conv-Reluplex framework}
%   \label{Figure 4.1}
% \end{figure}

\begin{itemize}
	\item Step 1: A correctly classified image sample is selected from the training dataset and input into the CNN.

	\item Step 2: When the image sample passes through the convolutional layer and the pooling layer, the rough features and the dominant features extracted by these two layers are stored respectively before it passes through the fully connected layer.

	\item Step 3: The stored dominant features are regarded as the input of a DNN, because the structure of the DNN is almost same as the fully connected layer. A small perturbation neighbourhood $\delta$ of the stored dominant features is a set, and   Reluplex is used to verify the adversarial robustness of the fully connected layer.

	\item Step 4: The adversarial robustness of the DNN (the fully connected layer) is verified from Reluplex. If it is not satisfied, Reluplex will output a counter-example to prove that the adversarial robustness is not satisfied. This example is called as "intermediate adversarial example".

	\item Step 5: The Unpooling algorithm (See Sec.III.A)  is applied to restore the intermediate adversarial example to the potential rough features before the  pooling operation.

	\item Step 6: The Deconvolution algorithm (See Sec.III.B) is applied to restore the potential rough features to a image.
\end{itemize}

Reluplex is able to find out a counter-example (the intermediate adversarial example in our framework), in general, Conv-reluplex can turn this intermediate adversarial example into a real image.

\subsection{Unpooling algorithm}
The function of pooling layers is to reduce the dimension of rough features and generate dominant features. The commonly used types of pooling calculation are Max, Average, Sum, etc.

\begin{figure}[htpb]
  \centering
  \includegraphics[width=5cm]{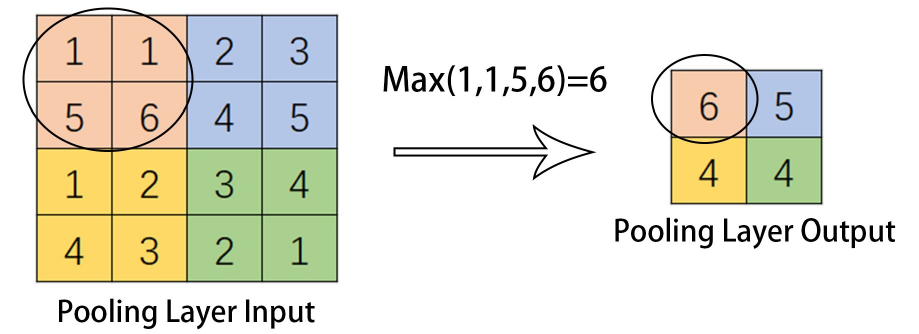}
  \caption{Max-pooling calculation}
  \label{Figure 4.2}
\end{figure}

The Max-pooling is shown in Fig. \ref{Figure 4.2}. The input of this pooling layer is a 4\(* 4\) matrix , the pooling window is a 2\(* 2\) matrix, and the stride is 2. The first iteration operation of the pooling layer is $Max (1,1,5,6)=6$, then sliding to the right of 2 cells, the second pooling operation is $Max (2,3,4,5)=5$, and so no. The final output of the pooling layer is a 2\(* 2\) matrix.

The Unpooling algorithm is the inverse calculation of pooling operation. The potential rough features can be generated based on the Unpooling algorithm using the intermediate adversarial example found in Step 4 and the original rough features stored in Step 3. Because the CNN is well-trained, the internal structure and weight parameters is fixed. Our approach is to make minimal changes to the original rough features so that it match the intermediate adversarial example.

For the Max-pooling, our Unpooling algorithm goes through every value in the pooling window. If the value is greater than the corresponding value in the intermediate adversarial example, the value is set as the corresponding value. If the value is less than or equal to its corresponding value, the value will remain. Assume that $x$ is the value in the original rough features, and \(\alpha\) is its corresponding value in the intermediate adversarial example, the Unpooling algorithm is defined as following:

\begin{equation*}
f_{unpooling}(x)=\left\{\begin{array}{ll}{x=\alpha,} & {\text { if\quad } x>\alpha} \\ {x=x,} & {\text { if\quad } x \leq \alpha}\end{array}\right.
\end{equation*}

\begin{figure}[htpb]
  \centering
  \includegraphics[width=8cm]{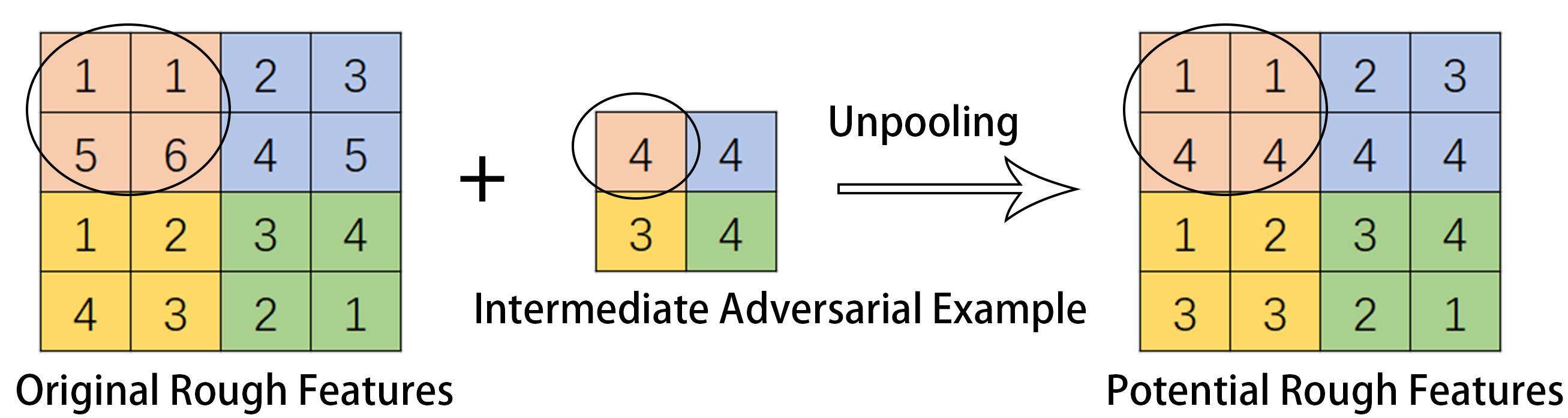}
  \caption{Unpooling calculation of Max-pooling}
  \label{Figure 4.3}
\end{figure}

An example of the Unpooling algorithm for the $Max-pooling$ is illustrated as Fig. \ref{Figure 4.3}. The left matrix is the original rough features, the middle matrix is the intermediate adversarial example. Firstly, the values 1,1,5,6 are in the window of the first pooling operation, and  compared with the corresponding value in the intermediate adversarial example 4. Secondly, because 5 and 6 are larger than 4, so the valuse of 5 and 6 reduce to 4, and other two values, 1 and 1, are both smaller than 4, so they remain. The pooling window slides to the next position iteratively. Finally, the new input matrix on the right side of Fig. \ref{Figure 4.3} is generated.

As mentioned before, the Unpooling algorithm only modifies the value necessarily. Therefore, the minimal changes can be ensured.

%----------------------------------------------------------

\subsection{Deconvolution algorithm}

\begin{figure}[htpb]
  \centering
  \includegraphics[width=6cm]{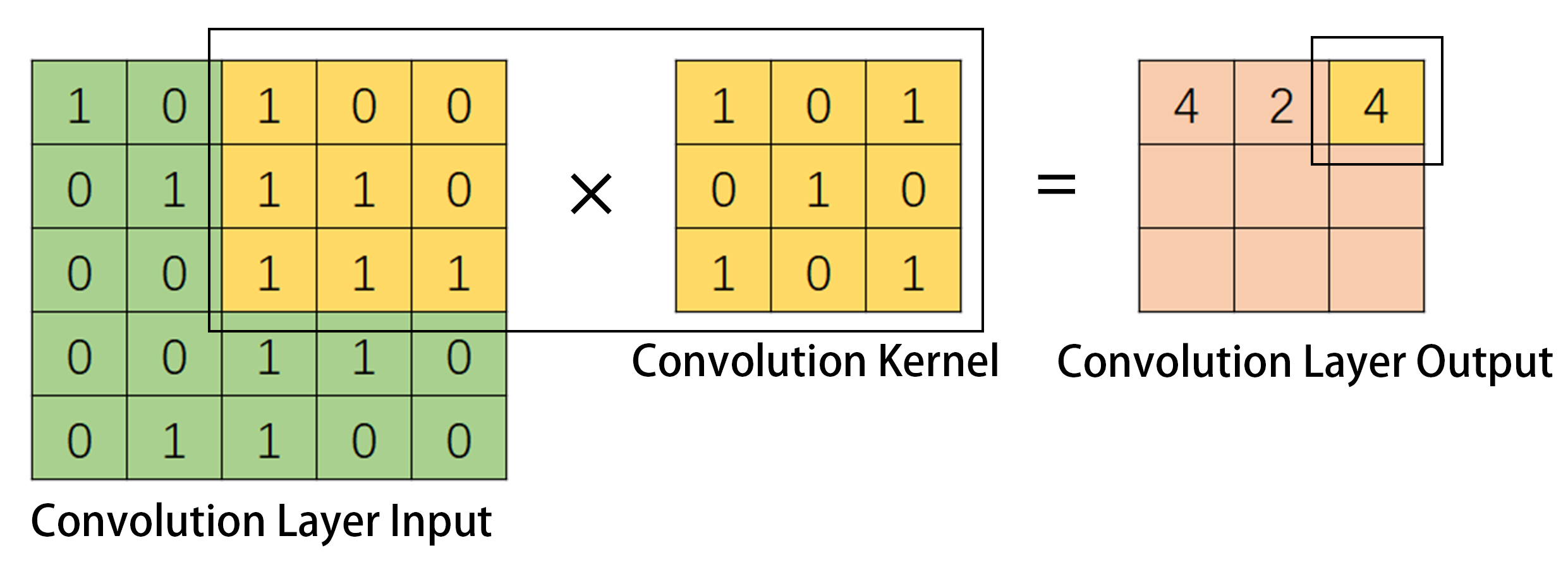}
  \caption{Convolution calculation}
  \label{Figure 4.4}
\end{figure}

The function of convolutional layers in CNNs is to extract the rough features from the original image, and the operation is demonstrated in Fig. \ref{Figure 4.4}. The input of the convolutional layer is a 5$*$5  matrix, the kernel (parameter) of this layer is a 3\(* 3\) matrix, and the stride is 1. When the third iteration of convolution is performed, The kernel slides to the upper right corner of the input. The corresponding values in the same location of these two matrices multiply with each other and the values of iterations are summed up. The result is shown at the upper right corner of the output matrix. When the operations of convolution are completed, the output of the convolutional layers, a 3\(* 3\) matrix, is obtained.

Generally, a nonlinear activation function follows the convolutional layer. ReLU function is a widely used nonlinear activation function after the convolutional layer, and it can be expressed as follows.
\begin{equation*}
Y_{i}=\operatorname{ReLU}\left(X_{i} \cdot W_{i}+B_{i}\right)
\end{equation*}
where \(X_{i}\) represents input data of layer \(i\), \(W_{i}\) represents convolution kernel parameter of layer \(i\), \(B_{i}\) represents bias of layer \(i\), and \(Y_{i}\) represents feature map of layer \(i\).

In order to convert the rough features back to the image which violate the adversarial robustness, the deconvolution algorithm should invert both the ReLU operation and the convolution operation. %According to the expression $Y_{i}=\operatorname{ReLU}\left(X_{i} \cdot W_{i}+B_{i}\right)$,
The problem of convolution layer inversion is formulated as a linear constraint solving problem. The kernel \(W_{i}\), the bias \(B_{i}\) and the potential rough features \(Y_{i}\) is obtained from Step 2 and Step 5 respectively. \(X_{i}\) is %a desired generated counter image
the adversarial image to be generated. Since the result of the ReLU function \(Y_{i}\) is known, according to \(Y_{i}\) and the ReLU definition, the ReLU constraint can be eliminated from the expression, and the problem is directly encoded as the following constraints:

\begin{footnotesize}
\begin{equation*}
Y_{i}=\operatorname{ReLU}\left(X_{i} \cdot W_{i}+B_{i}\right) \Leftrightarrow \left\{\begin{array}{l}{X_{i} \cdot W_{i}+B_{i}=Y_{i}, \quad \text {if \quad } Y_{i}>0} \\ {X_{i} \cdot W_{i}+B_{i} \leq 0, \quad \text { if \quad } Y_{i}=0}\end{array}\right.
\label{eq:de-conv}
\end{equation*}
\end{footnotesize}

For the first convolution layer, and the input of this layer is still the input of the CNN, an additional upper and lower bound constraint of \(0 \leq X_{i} \leq 1\) needs to be added for each variable \(X_{i}\). Because in preprocessing stage, the pixel values of 0 to 255 are usually normalised between 0 to 1, the additional constraints ensure that the generated data obtained by the inversion can be correctly converted into a image.

\begin{figure}[htpb]
  \centering
  \includegraphics[width=5cm]{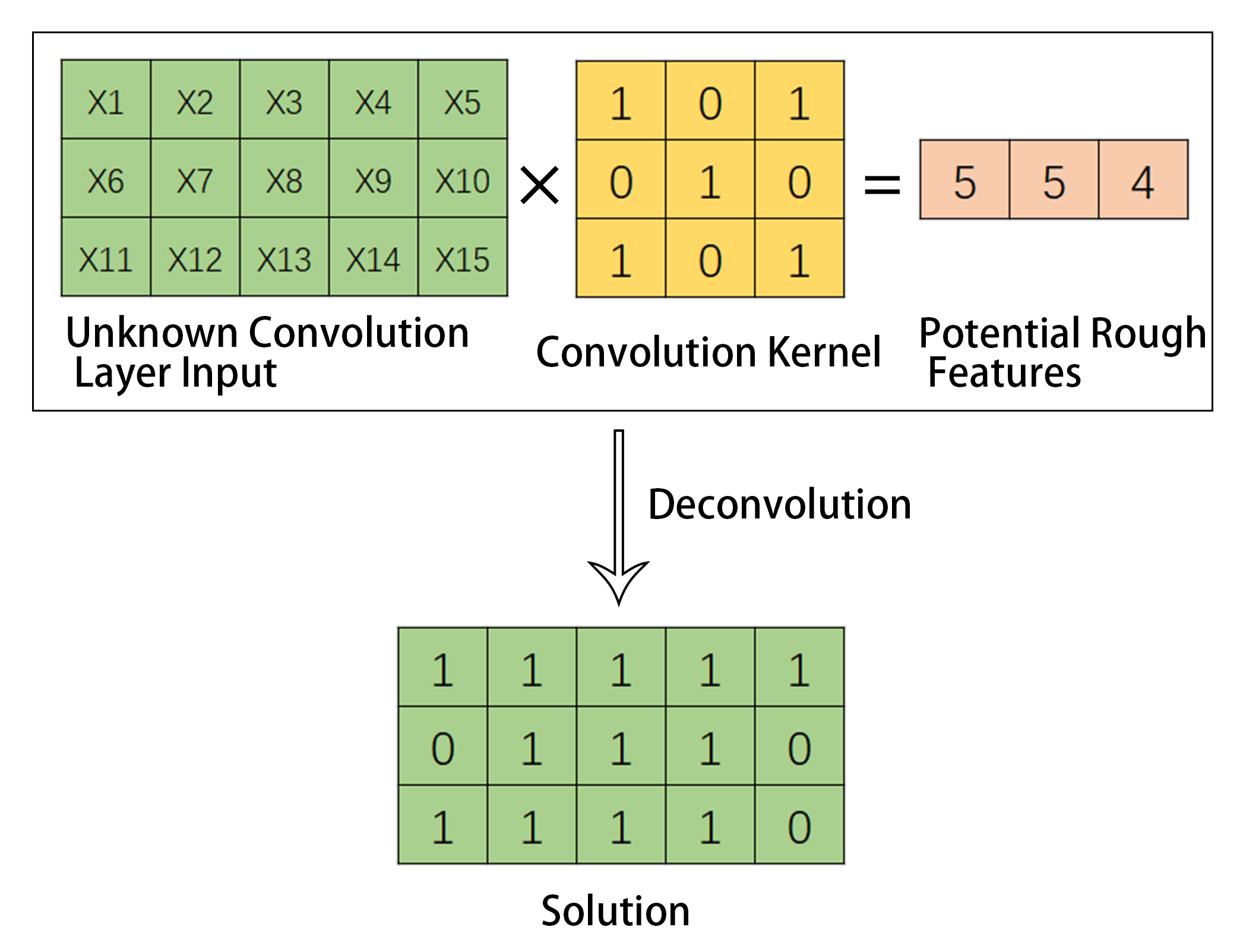}
  \caption{Deconvolution calculation}
  \label{Figure 4.5}
\end{figure}

An example of the Deconvolution algorithm is shown in Fig. \ref{Figure 4.5}. The input data (the normalised image) to be generated is a 3\(* 5\) matrix, and all values, \(x_{1}\),...,\(x_{15}\), in the matrix are unknown. The convolution kernel is a 3\(* 3\) matrix, the values of this kernel are already known, the bias are all 0 for simplicity, and the potential rough features is also known. Assuming that this is the first convolution layer, the inequality groups can be formulated as follows:

\begin{equation*}
\left\{\begin{array}{c}{x_{1}+x_{3}+x_{7}+x_{11}+x_{13}=5} \\ {x_{2}+x_{4}+x_{8}+x_{12}+x_{14}=5} \\ {x_{3}+x_{5}+x_{9}+x_{13}+x_{15}=4} \\ {0 \leq x_{i} \leq 1}\end{array}\right.
\end{equation*}

The solutions of these constraints are generally not unique, one of which is:
\begin{equation*}
\left\{\begin{array}{c}{x_{1}=1, x_{2}=1, x_{3}=1, x_{4}=1, x_{5}=1} \\ {x_{6}=0, x_{7}=1, x_{8}=1, x_{9}=1, x_{10}=0} \\ {x_{11}=1, x_{12}=1, x_{13}=1, x_{14}=1, x_{15}=0}\end{array}\right.
\end{equation*}

In practice, it is often the case that the constraints admit multiple solutions.
%If the number of the solutions is small, our algorithm generates several solutions.
If the number of the solutions is too large, our algorithm will further to solve an optimization problem (linear programming), taking the maximum or minimum value of the sum of all variables \(x_{i}\) as the objective function. The optimum can be directly selected as  a solution.

There are various mature algorithms and tools for solving  linear programs. The tool PuLP \cite{pulp} which is based on simplex algorithm is employed in our experiments. If the solution fails, it means that there are obvious conflicts between the constraints, and the deconvolution algorithm can not be continued. In this circumstance, the corresponding ``intermediate adversarial examples'' is considered as spurious and will be discarded. If the solution can be found, it means that the decalculation of the current convolutional layer is successful. If the previous layer is another pooling layer, the Unpooling algorithm is used to process again. Otherwise, a couter-example image might be generated.

%====================================================================================================

\section{Experimental Results}

\subsection{Conv-Reluplex  experiments }
This experiment includes two steps. The first step is to train a CNN model. The second step is to verify the CNN model by using the Conv-Reluplex framework and generate adversarial examples, which demonstrates the effectiveness of our Conv-Reluplex algorithm.

\subsubsection{Training a CNN model}

\begin{figure*}[htpb]
  \centering
  \includegraphics[scale=0.45]{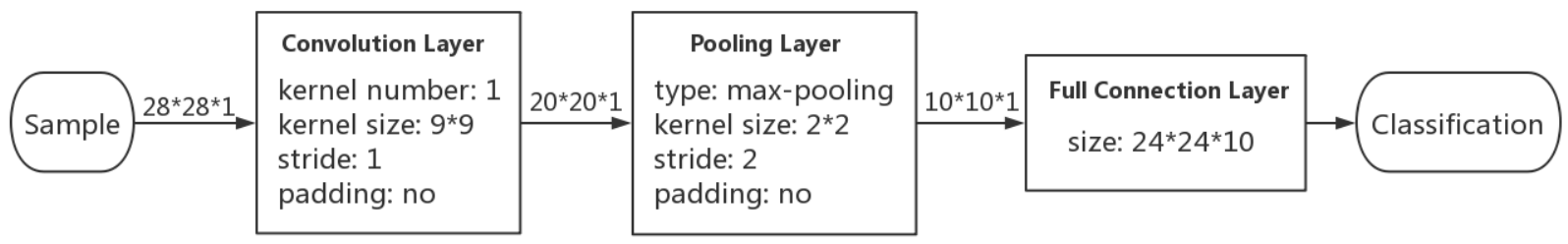}
  \caption{The structure of the CNN model in the experiment}
  \label{Figure 5.1}
\end{figure*}

The public MNIST data set \cite{lecun1998mnist} is employed for this experiment. MNIST is an image data set of handwritten numbers 0 to 9, including 60,000 training samples and 10,000 test samples. Each sample is a grayscale image with the size of 28\(* 28\) and the grayscale range of 0 to 255. In preprocessing, the grayscale is normalized as data between 0 and 1. The structure of the CNN model we built is shown in Fig. \ref{Figure 5.1} with batch size 64 and   epoch 10. After the training step, the test loss is 0.1894 and the test accuracy is 94.98\%.

\subsubsection{Verifying and generating picture adversarial examples by Conv-Reluplex}

An image sample of handwritten numeral 6 is used to verify adversarial robustness. We input the image sample into the CNN model, extract feature data at the full connection layer, and set perturbation neighborhood \(\delta\) from 0.2 to 0.6. (Note: the perturbation neighborhood of 0.2 means that each generated value is between the original value adding 0.2 and substrating 0.2. For example, the original input is [1.0, 1.5], and the perturbation [0.8, 1.3], [1.2, 1.7] both satisfy the setting of the perturbation neighborhood of 0.2, but [1.0, 1.8] is under the 0.3 perturbation neighborhood rather than 0.2. The verification results are shown in Table. \ref{Verification results of a test sample of number 6 in different perturbation neighborhoods}:

\begin{table}[htpb]
\centering
\caption{Verification results of a test sample of number 6 in different perturbation neighborhoods}
\label{Verification results of a test sample of number 6 in different perturbation neighborhoods}

\resizebox{254pt}{46pt}{
\begin{tabular}{|c|c|c|c|c|c|c|c|c|c|c|}
\hline
 \diagbox[width=100pt, height=26pt]{perturbation\\neighborhood }{classification   }& 0       & 1       & 2       & 3       & 4       & 5       & 6 & 7       & 8       & 9       \\ \hline
$\delta$ = 0.2   & $\surd$ & $\surd$ & $\surd$ & $\surd$ & $\surd$ & $\surd$ & $\backslash$ & $\surd$ & $\surd$ & $\surd$ \\ \hline
$\delta$ = 0.3   & $\surd$ & $\surd$ & $\surd$ & $\surd$ & $\surd$ & Fail    & $\backslash$ & $\surd$ & $\surd$ & $\surd$ \\ \hline
$\delta$ = 0.4   & $\surd$ & $\surd$ & $\surd$ & $\surd$ & $\surd$ & Fail    & $\backslash$ & $\surd$ & $\surd$ & $\surd$ \\ \hline
$\delta$ = 0.5   & $\surd$ & $\surd$ & $\surd$ & $\surd$ & $\surd$ & Fail    & $\backslash$ & $\surd$ & Fail    & $\surd$ \\ \hline
$\delta$ = 0.6   & $\surd$ & $\surd$ & $\surd$ & $\surd$ & $\surd$ & Fail    & $\backslash$ & $\surd$ & Fail    & $\surd$ \\ \hline
\end{tabular}}
\end{table}

The symbol "$\surd$" indicates that there is no adversarial example here. "$\backslash$" indicates that the classification is the real class of the sample, and verification can be skipped.  ``Fail'' means that there is at least one adversarial example misclassified as the corresponding class in the specified perturbation neighborhood. Only when there is no ``Fail'' in an entire row, the adversarial robustness in the corresponding neighborhood is satisfied.

As shown in Table \ref{Verification results of a test sample of number 6 in different perturbation neighborhoods}, when $\delta$ = 0.2, the adversarial robustness is satisfied, and no other classification will appear. When $\delta$ = 0.3, the adversarial robustness is not satisfied, and the sample of number 6 will be classified as number 5.  As the neighborhood $\delta$ increases, the misclassification caused by perturbation becomes more and more serious. Such as $\delta$ = 0.5 or $\delta$ = 0.6, the selected sample of number 6 will be classified as numbers 5 or 8. It is obvious that the number of misclassification types and  the number of intermediate adversarial examples have further increased.

\begin{figure}[htpb]
  \centering
  \includegraphics[width=5cm]{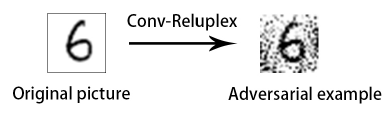}
  \caption{Original picture and adversarial example of number 6}
  \label{Figure 5.2}
\end{figure}

An intermediate adversarial example misclassified as 5 is chosen when \(\delta\) = 0.6 to carry out further analysis. The intermediate adversarial example is converted to an image by using the Unpooling and Deconvolution algorithm of Conv-Reluplex. The comparison between the original image and the adversarial example image is shown in Fig. \ref{Figure 5.2}. Although there is a lot of noise in the adversarial example picture, it does not prevent human from recognizing the main content. %It is still easy to recognize the number 6, but the CNN model predicts it to be the number 5.

In addition to the number 6, other numbers are also selected, and  the different types of original samples are used to generate different adversarial examples, as shown in Fig. \ref{Figure 5.3}:
\begin{figure}[htpb]
  \centering
  \includegraphics[width=7cm]{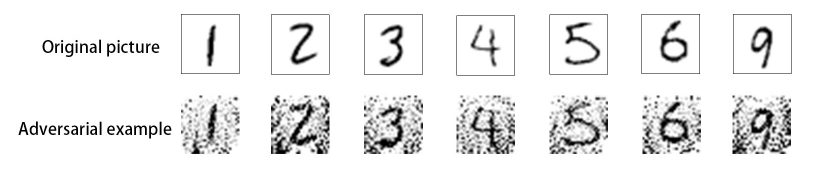}
  \caption{Different adversarial examples}
  \label{Figure 5.3}
\end{figure}

\subsection{Adversarial training experiment}
This experiment is designed to enhance the robustness of the model by adversarial training. The adversarial examples used in training are all generated by Conv-Reluplex. The numbers 1, 3 and 6 are selected as the main original samples. Each number has generated 200 adversarial examples, of which 100 are used for adversarial training and the remaining 100 are used for testing. In order to prevent the model from forgetting the characteristics of the original samples in adversarial training, we randomly selected 900 or 1000 samples from the original data set of each number, and mixed them with adversarial examples. Therefore, in the new training set, the numbers 1, 3 and 6 are composed of 100 adversarial examples and 900 original samples. The remaining numbers are composed of 1000 original samples respectively. After preparing the new training set, the adversarial training is performed based on the original model, without changing any parameters or structure. The batch size is 64 and the epoch is 10.

The difference of the accuracy after adversarial training is shown in Table \ref{Comparison of accuracy of 10,000 samples before and after training}. The classification accuracy of adversarial examples has suddenly increased from 0\% to 99.65\%, while the accuracy of original samples has only decreased from 94.98\% to 94.89\%, and the difference is only 0.09\%. It shows that appropriate adversarial training can enhance the adversarial robustness of model, at the same time, the accuracy of original samples will not be affected too much.

\begin{table}[htpb]
\centering
\caption{Accuracy comparison before and after adversarial training}
\label{Comparison of accuracy of 10,000 samples before and after training}
\resizebox{88mm}{7mm}{
\begin{tabular}{|c|c|c|}
\hline
                          & accuracy of original sample  & accuracy of adversarial sample \\ \hline
Original  model      & 94.98\%                     & 0\%                             \\ \hline
Adversaria training  & 94.89\%                     & 99.65\%                         \\ \hline
\end{tabular}}
\end{table}

In order to further analyze the difference of the model's adversarial robustness after adversarial training, we respectively select one original sample of numbers 1, 3 and 6 as test data, and input them into the original model and the adversarial training model, then use Reluplex to verify the adversarial robustness of CNN's full connection layer. The verification results are shown in Table \ref{Verification results of test samples of number 1 in different models}-\ref{Verification results of test samples of number 6 in different models}.

\begin{table}[htpb]
\centering
\caption{Verification results of number 1's test sample \((\delta=0.6)\)}
\label{Verification results of test samples of number 1 in different models}
\resizebox{88mm}{7mm}{
\begin{tabular}{|c|c|c|c|c|c|c|c|c|c|c|}
\hline
classification            & 0       & 1 & 2       & 3       & 4       & 5       & 6       & 7    & 8       & 9    \\ \hline
Original model       & $\surd$ & $\backslash$ & $\surd$ & Fail    & Fail    & $\surd$ & $\surd$ & Fail & Fail    & Fail \\ \hline
Adversaria training  & $\surd$ & $\backslash$ & $\surd$ & $\surd$ & $\surd$ & $\surd$ & $\surd$ & Fail & $\surd$ & Fail \\ \hline
\end{tabular}}
\end{table}

\begin{table}[htpb]
\centering
\caption{Verification results of number 3's test sample \((\delta=0.6)\)}
\label{Verification results of test samples of number 3 in different models}
\resizebox{88mm}{7mm}{
\begin{tabular}{|c|c|c|c|c|c|c|c|c|c|c|}
\hline
classification            & 0       & 1       & 2       & 3 & 4       & 5       & 6       & 7       & 8       & 9       \\ \hline
Original model       & $\surd$ & $\surd$ & $\surd$ & $\backslash$ & $\surd$ & Fail    & $\surd$ & $\surd$ & Fail    & $\surd$ \\ \hline
Adversaria training  & $\surd$ & $\surd$ & $\surd$ & $\backslash$ & $\surd$ & $\surd$ & $\surd$ & $\surd$ & $\surd$ & $\surd$ \\ \hline
\end{tabular}}
\end{table}

\begin{table}[htpb]
\centering
\caption{Verification results of number 6's test sample \((\delta=0.6)\)}
\label{Verification results of test samples of number 6 in different models}
\resizebox{88mm}{7mm}{
\begin{tabular}{|c|c|c|c|c|c|c|c|c|c|c|}
\hline
classification      & 0       & 1       & 2       & 3       & 4       & 5       & 6 & 7       & 8       & 9       \\ \hline
Original model      & $\surd$ & $\surd$ & $\surd$ & $\surd$ & $\surd$ & Fail    & $\backslash$ & $\surd$ & Fail    & $\surd$ \\ \hline
Adversaria training & $\surd$ & $\surd$ & $\surd$ & $\surd$ & $\surd$ & $\surd$ & $\backslash$ & $\surd$ & $\surd$ & $\surd$ \\ \hline
\end{tabular}}
\end{table}

In Table \ref{Verification results of test samples of number 1 in different models}, when perturbation neighborhood \(\delta\) = 0.6, the selected sample of number 1 has five kinds of adversarial examples in original model, the selected sample may be misclassified as 3, 4, 7, 8 or 9. However, in the adversarial training model, the selected sample only has two kinds of adversarial examples. It may be misclassified as 7 or 9. We can find that the types of adversarial examples have decreased.

In Table \ref{Verification results of test samples of number 3 in different models}, the selected sample of number 3 has two kinds of adversarial examples in original model, which may be misclassified as 5 or 8. But in the adversarial training model, the two misclassification cases have disappeared, the selected sample has no adversarial examples in the same neighborhood \((\delta=0.6)\), and the adversarial robustness has changed from not satisfied to satisfied.

In Table \ref{Verification results of test samples of number 6 in different models}, the verification situation of number 6's selected sample is the same as number 3. It shows that this phenomenon is not accidental. Adversarial training can indeed repair the weakness of the model and enhance the adversarial robustness.

\section{Conclusion}
%In this paper, we propose a verification framework for CNN, called Conv-Reluplex. The Conv-Reluplex method logically regards the full connection layer as one part and the convolution layer and pooling layer as another part, i.e. the full connection layer is regarded as a single small DNNs to verify and generate intermediate layer adversarial examples, which are directly processed by the Reluplex algorithm.  As another part, the convolution layer and pooling layer are processed by our extended Unpooling algorithm and  Deconvolution algorithm.  In this way, the generated intermediate adversarial examples can be reversely deduced until it is restored to an image.  The above is the key point in the derivation process of Conv-Reluplex verification framework, which enables us to verify the safety of the convolution neural networks.

%In this paper, we propose a verification framework for convolution neural networks, called Conv-Reluplex.  The counter-example found by the Reluplex is extended to generate the adversarial images, which can be used to verify the adversarial robustness of CNNs.
In this paper, to make Reluplex more practical, we propose a Conv-Reluplex verification framework, which is utilized to check adversarial robustness of CNNs. In case the robustness property is not satisfied, it generates adversarial example. Using these adversarial examples to proceed adversarial training,  can indeed enhance the adversarial robustness of our model.

There are still some further work needed to be done.
The complexity of the general SMT  solver algorithm is exponential. Reluplex proposed its own optimization algorithm to solve some performance bottlenecks, but the efficiency is severe limited by the high nonlinearity of the resulting formulas. So we can only deal with some smaller networks at present. The network trained by MNIST only has more than 4,000 relu nodes, while the network nodes trained by ImageNet are too many to handle. We hope to do some work to improve efficiency so that it can handle more network nodes, and it is one of our goals to handle color pictures.
Finally, the recurrent neural networks (RNNs) can be taken into the consideration in both original Reluplex and the extensions of Reluplex.

% \section*{Acknowledgment}
%
% The preferred spelling of the word ``acknowledgment'' in America is without
% an ``e'' after the ``g''. Avoid the stilted expression ``one of us (R. B.
% G.) thanks $\ldots$''. Instead, try ``R. B. G. thanks$\ldots$''. Put sponsor
% acknowledgments in the unnumbered footnote on the first page.
%

\section*{Acknowledgment}
 We acknowledge the support of  National Natural Science Foundation of China (NSFC) Project 61972284.

\bibliographystyle{IEEEtran}
\bibliography{reference}

\begin{thebibliography}{10}
\providecommand{\url}[1]{#1}
\csname url@samestyle\endcsname
\providecommand{\newblock}{\relax}
\providecommand{\bibinfo}[2]{#2}
\providecommand{\BIBentrySTDinterwordspacing}{\spaceskip=0pt\relax}
\providecommand{\BIBentryALTinterwordstretchfactor}{4}
\providecommand{\BIBentryALTinterwordspacing}{\spaceskip=\fontdimen2\font plus
\BIBentryALTinterwordstretchfactor\fontdimen3\font minus
  \fontdimen4\font\relax}
\providecommand{\BIBforeignlanguage}[2]{{%
\expandafter\ifx\csname l@#1\endcsname\relax
\typeout{** WARNING: IEEEtran.bst: No hyphenation pattern has been}%
\typeout{** loaded for the language `#1'. Using the pattern for}%
\typeout{** the default language instead.}%
\else
\language=\csname l@#1\endcsname
\fi
#2}}
\providecommand{\BIBdecl}{\relax}
\BIBdecl

\bibitem{goodfellow2016deep}
I.~Goodfellow, Y.~Bengio, and A.~Courville, \emph{Deep learning}.\hskip 1em
  plus 0.5em minus 0.4em\relax MIT press, 2016.

\bibitem{riesenhuber1999hierarchical}
M.~Riesenhuber and T.~Poggio, ``Hierarchical models of object recognition in
  cortex,'' \emph{Nature neuroscience}, vol.~2, no.~11, p. 1019, 1999.

\bibitem{krizhevsky2012imagenet}
A.~Krizhevsky, I.~Sutskever, and G.~E. Hinton, ``Imagenet classification with
  deep convolutional neural networks,'' in \emph{Advances in neural information
  processing systems}, 2012, pp. 1097--1105.

\bibitem{graves2013speech}
A.~Graves, A.-r. Mohamed, and G.~Hinton, ``Speech recognition with deep
  recurrent neural networks,'' in \emph{2013 IEEE international conference on
  acoustics, speech and signal processing}.\hskip 1em plus 0.5em minus
  0.4em\relax IEEE, 2013, pp. 6645--6649.

\bibitem{zhao2017deep}
D.~Zhao, Y.~Chen, and L.~Lv, ``Deep reinforcement learning with visual
  attention for vehicle classification,'' \emph{IEEE Transactions on Cognitive
  and Developmental Systems}, vol.~9, no.~4, pp. 356--367, 2017.

\bibitem{szegedy2013intriguing}
C.~Szegedy, W.~Zaremba, I.~Sutskever, J.~Bruna, D.~Erhan, I.~Goodfellow, and
  R.~Fergus, ``Intriguing properties of neural networks,'' \emph{arXiv preprint
  arXiv:1312.6199}, 2013.

\bibitem{pulina2012challenging}
L.~Pulina and A.~Tacchella, ``Challenging smt solvers to verify neural
  networks,'' \emph{Ai Communications}, vol.~25, no.~2, pp. 117--135, 2012.

\bibitem{bastani2016measuring}
O.~Bastani, Y.~Ioannou, L.~Lampropoulos, D.~Vytiniotis, A.~Nori, and
  A.~Criminisi, ``Measuring neural net robustness with constraints,'' in
  \emph{Advances in neural information processing systems}, 2016, pp.
  2613--2621.

\bibitem{huang2017safety}
X.~Huang, M.~Kwiatkowska, S.~Wang, and M.~Wu, ``Safety verification of deep
  neural networks,'' in \emph{International Conference on Computer Aided
  Verification}.\hskip 1em plus 0.5em minus 0.4em\relax Springer, 2017, pp.
  3--29.

\bibitem{pulina2010abstraction}
L.~Pulina and A.~Tacchella, ``An abstraction-refinement approach to
  verification of artificial neural networks,'' in \emph{International
  Conference on Computer Aided Verification}.\hskip 1em plus 0.5em minus
  0.4em\relax Springer, 2010, pp. 243--257.

\bibitem{rumelhart1985learning}
D.~E. Rumelhart, G.~E. Hinton, and R.~J. Williams, ``Learning internal
  representations by error propagation,'' California Univ San Diego La Jolla
  Inst for Cognitive Science, Tech. Rep., 1985.

\bibitem{katz2017reluplex}
G.~Katz, C.~Barrett, D.~L. Dill, K.~Julian, and M.~J. Kochenderfer, ``Reluplex:
  An efficient smt solver for verifying deep neural networks,'' in
  \emph{International Conference on Computer Aided Verification}.\hskip 1em
  plus 0.5em minus 0.4em\relax Springer, 2017, pp. 97--117.

\bibitem{goodfellow2014explaining}
I.~J. Goodfellow, J.~Shlens, and C.~Szegedy, ``Explaining and harnessing
  adversarial examples,'' \emph{arXiv preprint arXiv:1412.6572}, 2014.

\bibitem{moosavi2016deepfool}
S.-M. Moosavi-Dezfooli, A.~Fawzi, and P.~Frossard, ``Deepfool: a simple and
  accurate method to fool deep neural networks,'' in \emph{Proceedings of the
  IEEE conference on computer vision and pattern recognition}, 2016, pp.
  2574--2582.

\bibitem{brendel2017decision}
W.~Brendel, J.~Rauber, and M.~Bethge, ``Decision-based adversarial attacks:
  Reliable attacks against black-box machine learning models,'' \emph{arXiv
  preprint arXiv:1712.04248}, 2017.

\bibitem{katz2017towards}
G.~Katz, C.~Barrett, D.~L. Dill, K.~Julian, and M.~J. Kochenderfer, ``Towards
  proving the adversarial robustness of deep neural networks,'' \emph{arXiv
  preprint arXiv:1709.02802}, 2017.

\bibitem{Vanderbei1998Linear}
R.~J. Vanderbei, ``Linear programming: Foundations and extensions,''
  \emph{Journal of the Operational Research Society}, vol.~49, no.~1, pp.
  94--94, 1998.

\bibitem{dantzig2016linear}
G.~Dantzig, \emph{Linear programming and extensions}.\hskip 1em plus 0.5em
  minus 0.4em\relax Princeton university press, 2016.

\bibitem{cnndefinition}
\url{http://ufldl.stanford.edu/tutorial/supervised/ConvolutionalNeuralNetwork/}.

\bibitem{julian2018deep}
K.~D. Julian, M.~J. Kochenderfer, and M.~P. Owen, ``Deep neural network
  compression for aircraft collision avoidance systems,'' \emph{Journal of
  Guidance, Control, and Dynamics}, vol.~42, no.~3, pp. 598--608, 2018.

\bibitem{pulp}
\url{https://pythonhosted.org/PuLP/}.

\bibitem{lecun1998mnist}
Y.~LeCun, ``The mnist database of handwritten digits,'' \emph{http://yann.
  lecun. com/exdb/mnist/}, 1998.

\end{thebibliography}

\end{document}